\documentclass[11pt]{article}

\usepackage[final]{acl}

\usepackage{times}
\usepackage{latexsym}
\usepackage{amsmath}
\usepackage{amsthm}
\usepackage{amssymb}
\theoremstyle{definition}

\usepackage[T1]{fontenc}
\usepackage[utf8]{inputenc}
\usepackage{amsmath}
\usepackage{bbm}

\usepackage{microtype}
\usepackage{multirow}

\usepackage{inconsolata}

\usepackage{graphicx}
\usepackage{booktabs}
\usepackage{arydshln}

\newcommand{\rrlcf}{R}

\usepackage{amsmath, amssymb}
\usepackage{tcolorbox}
\tcbuselibrary{breakable}
\usepackage{listings}
\usepackage{cuted}
\usepackage{fancyvrb}
\title{The Confidence Dichotomy: Analyzing and Mitigating Miscalibration in Tool-Use Agents}

\author{
  Weihao Xuan$^{1,2}$\thanks{Equal contribution.}, Qingcheng Zeng$^{3}$\footnotemark[1], Heli Qi$^{2,4}$, Yunze Xiao$^{5}$, \\\textbf{Junjue Wang}$^{1}$\textbf{,} \textbf{Naoto Yokoya}$^{1,2}$\thanks{Corresponding author.} \\
  $^{1}$The University of Tokyo, $^{2}$RIKEN AIP, $^{3}$Northwestern University \\ $^{4}$Waseda University, $^{5}$Carnegie Mellon University \\
}

\begin{document}
\maketitle
\begin{abstract}
Autonomous agents based on large language models (LLMs) are rapidly evolving to handle multi-turn tasks, but ensuring their trustworthiness remains a critical challenge. A fundamental pillar of this trustworthiness is calibration, which refers to an agent's ability to express confidence that reliably reflects its actual performance. While calibration is well-established for static models, its dynamics in tool-integrated agentic workflows remain under-explored. In this work, we systematically investigate verbalized calibration in tool-use agents, revealing a fundamental confidence dichotomy driven by tool type. Specifically, our pilot study identifies that evidence tools (e.g., web search) systematically induce severe overconfidence due to inherent noise in retrieved information, while verification tools (e.g., code interpreters) can ground reasoning through deterministic feedback and mitigate miscalibration. To robustly improve calibration across tool types, we propose a reinforcement learning (RL) fine-tuning framework that jointly optimizes task accuracy and calibration, supported by a holistic benchmark of reward designs. We demonstrate that our trained agents not only achieve superior calibration but also exhibit robust generalization from local training environments to noisy web settings and to distinct domains such as mathematical reasoning. Our results highlight the necessity of domain-specific calibration strategies for tool-use agents. More broadly, this work establishes a foundation for building self-aware agents that can reliably communicate uncertainty in high-stakes, real-world deployments.
\end{abstract}

\section{Introduction}
Autonomous agents based on LLMs represent a transformative leap in artificial intelligence, evolving beyond static text processing to engage with dynamic, real-world environments actively. By leveraging external tools and iterative reasoning, these systems are demonstrating rapid increases in proficiency in executing complex, long-horizon tasks that were previously intractable. The research community has recently witnessed remarkable strides across critical domains, including sophisticated code agents \cite{jimenez2024swebenchlanguagemodelsresolve, yang2024sweagent, yang2025swesmithscalingdatasoftware, traeresearchteam2025traeagentllmbasedagent}, autonomous web agents \cite{he2024webvoyager, wei2025webagentr1trainingwebagents}, and advanced deep search agents \cite{wu2025webdancerautonomousinformationseeking, li2025websailornavigatingsuperhumanreasoning, li2025websailorv2bridgingchasmproprietary, tongyidr}. The implications of this progress extend far beyond technical novelty. Indeed, these agents are poised to exert a substantial economic impact and fundamentally reshape the future of work \cite{shao2025futureworkaiagents}.

While the trustworthiness and safety of LLMs have been examined extensively, research specifically addressing trustworthiness in multi-turn agents remains sparse \cite{hua-etal-2024-trustagent, yu2025surveytrustworthyllmagents, shi2025trustworthyguiagentssurvey}. A fundamental pillar of this trustworthiness is \textit{calibration}, which refers to an agent's ability to report confidence scores that reliably reflect its actual performance. Within the limited existing literature, search agent-oriented benchmarks \cite{wei2025browsecompsimplechallengingbenchmark, zhou2025browsecompzhbenchmarkingwebbrowsing} consistently report that tool-use agents exhibit higher calibration errors than standalone LLMs, suggesting that external tools often exacerbate overconfidence. However, these studies focus narrowly on search scenarios, leaving open a critical question: \textit{is miscalibration a universal consequence of tool use, or does it depend on the nature of the tool itself?}

In this work, we present a systematic pilot study that answers this question by comparing representative tool-use agents \cite{jin2025searchr1trainingllmsreason, xue2025simpletirendtoendreinforcementlearning} against their standard instruction-following counterparts. Our analysis reveals a critical \textbf{confidence dichotomy}: not all tools affect calibration equally. Specifically, we identify two distinct tool categories with opposing effects. \textbf{Evidence tools} (e.g., web search), which retrieve external information laden with noise and uncertainty, systematically induce severe overconfidence. In contrast, \textbf{verification tools} (e.g., code interpreters), which provide deterministic execution feedback, can ground reasoning and mitigate miscalibration. This dichotomy persists across both prompting-based strategies and standard tool-use-oriented RL, indicating a fundamental challenge that cannot be resolved through prompt engineering alone.

To improve calibration across tool-use scenarios, we propose the \textbf{Calibration Agentic RL (CAR)} framework, a novel RL-based fine-tuning approach that jointly optimizes task accuracy and the reliability of expressed confidence. Through a holistic evaluation of diverse reward structures, we demonstrate that our trained agents maintain competitive accuracy while achieving significantly better calibration than baselines. Furthermore, we validate the robustness of our approach: agents trained in controlled local retriever environments generalize effectively to more noisy, API-based web search scenarios, and the framework proves effective in tool-integrated mathematical reasoning. Our primary contributions are summarized as follows:
\begin{itemize}
    \item We conduct a systematic pilot study that reveals a critical confidence dichotomy in tool-use agents: while verification tools provide grounding, evidence tools inherently predispose agents to severe overconfidence.
    \item We propose CAR, an RL-based framework for optimizing agent calibration, supported by a holistic benchmark of reward structures, including our novel Margin-Separated Calibration Reward (MSCR), that provides key insights for future reward design.
    \item We validate the effectiveness and robustness of our methodology across distinct task domains and demonstrate successful cross-environment generalization from local to noisy web settings.
\end{itemize}

\section{Related Work}
\paragraph{Calibration in LLMs} The calibration of LLMs has emerged as a central theme in recent literature. Verbalized confidence \cite{lin2022teachingmodelsexpressuncertainty,tian-etal-2023-just} has gained prominence due to its inherent interpretability and simplicity, with subsequent evaluations across instruct LLMs \cite{xiong2024llmsexpressuncertaintyempirical}, reasoning LLMs \cite{zeng-etal-2025-thinking, yoon2025reasoningmodelsbetterexpress}, and vision-language models \cite{xuan-etal-2025-seeing, liu2025taming} consistently characterizing these models as exhibiting moderate overconfidence. In terms of training, \citet{damani2025binaryrewardstraininglms} proposed an RL-based framework to encourage calibration in single-turn LLMs. Despite extensive investigation into static LLMs, calibration in autonomous agents remains notably sparse. Recent work proposes a post-hoc framework that trains an external predictor to estimate trajectory success~\citep{AgenticConfidenceCalibration}. However, this approach neither analyzes how different tool types systematically affect confidence dynamics nor improves the agent's intrinsic verbalized calibration. To address these gaps, we first conduct a systematic pilot study examining how different tool types affect agent calibration, then propose an RL-based framework to improve intrinsic verbalized confidence.

\paragraph{Tool-use Agents} The paradigm of LLMs is shifting from static text generation to autonomous agents capable of interacting with external environments. Equipping agents with tool-use capabilities enables them to overcome inherent limitations, such as hallucinations and calculation errors. Recent literature has explored diverse tool integration strategies, which can be broadly categorized by their function. On one hand, \textit{evidence tools} such as web search enable agents to retrieve external information to augment their knowledge; for instance, Search-R1 \cite{jin2025searchr1trainingllmsreason} leverages RL to intrinsically motivate agents to seek information when their internal knowledge is insufficient. On the other hand, \textit{verification tools} such as code interpreters provide deterministic feedback to validate reasoning steps. \citet{xue2025simpletirendtoendreinforcementlearning} integrates external Python interpreters to enhance agents' mathematical reasoning capabilities through robust code execution. However, despite these advancements making agents significantly more capable, emerging evidence suggests a critical side effect: tool-induced overconfidence. Preliminary results indicate that agents often place blind trust in tool outputs or overestimate their ability to solve tasks simply because tools are available \cite{wei2025browsecompsimplechallengingbenchmark, zhou2025browsecompzhbenchmarkingwebbrowsing}. In this paper, we systematically analyze this overconfidence across different tool types and propose mechanisms to mitigate it. 

\begin{figure*}[hbt]
  \centering
  \includegraphics[width=0.9\linewidth]{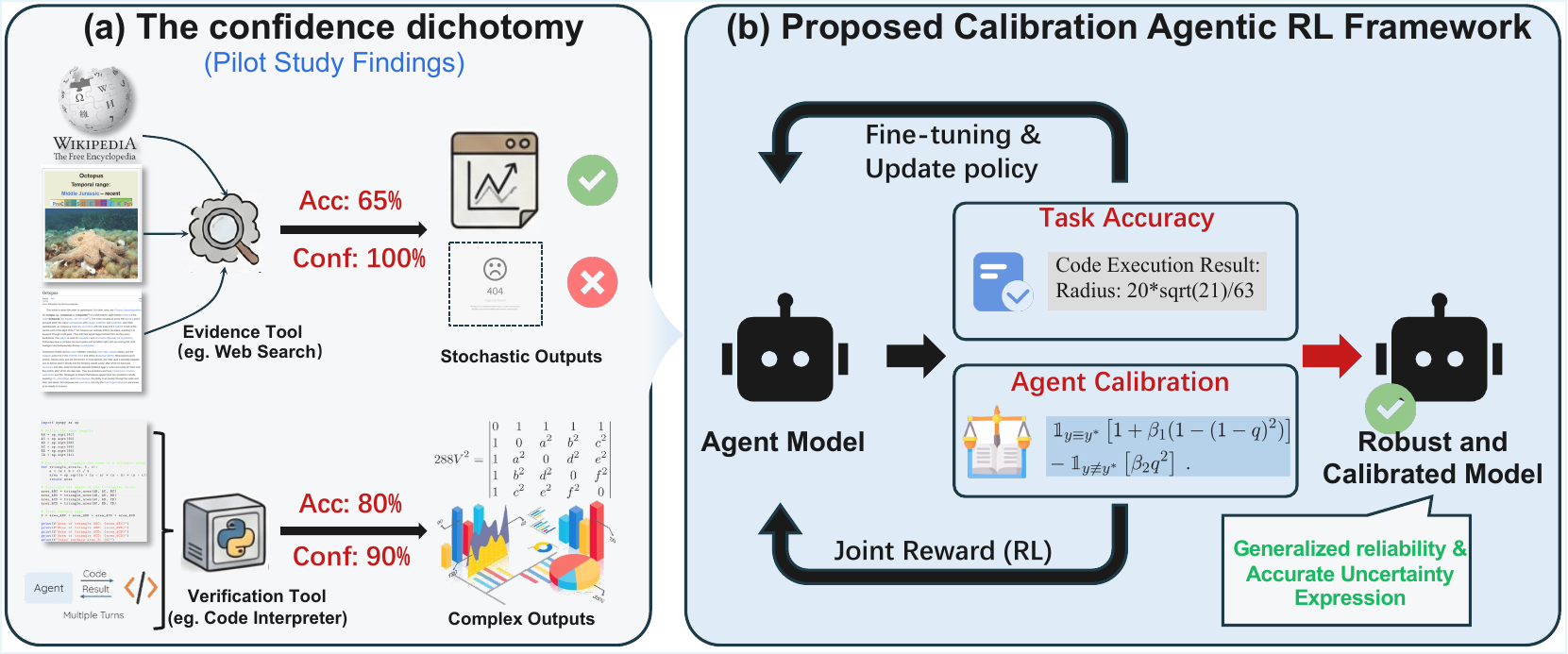}
  \caption{\textbf{The confidence dichotomy and the proposed RL framework.} (a) Our pilot study reveals a tool-dependent divergence in calibration: \textit{evidence tools} (e.g., web search), which operate in noisy retrieval environments, systematically induce overconfidence. In contrast, \textit{verification tools} (e.g., code interpreters), which provide deterministic execution feedback, exhibit better alignment between confidence and accuracy. (b) To address this miscalibration, we fine-tune agents with a joint RL objective that combines task accuracy and calibration rewards, producing robust agents with reliable uncertainty expression.}
\label{fig:framework}
\end{figure*}

\section{Pilot Study}
\label{sec:pilot_study}
\subsection{Overall Motivation}
The primary objective of this pilot study is to answer the question raised in our introduction: \textit{is miscalibration a universal consequence of tool use, or does it depend on the nature of the tool itself?} To this end, we systematically isolate and quantify the impact of different tool types on LLM calibration. We specifically analyze the confidence shifts that occur when transitioning from standard text generation to tool-augmented agentic workflows. By contrasting distinct operational modes, including standard prompting, prompting-based tool use, and RL-optimized tool use, we aim to elucidate how invoking external tools fundamentally modulates the reliability of model confidence.

\subsection{Experimental Setup}
To ensure a rigorous comparison, we establish three distinct experimental configurations:
\begin{enumerate}
    \item \textbf{Direct Prompting:} The LLM is instructed to address queries utilizing only its internal parametric knowledge.
    \item \textbf{Prompting-based Tool-Use:} The LLM is prompted to function as an autonomous agent with the capacity to invoke external tools, without updates to its model weights.
    \item \textbf{RL-based Tool-Use:} This setting adopts similar agentic prompts as the second configuration. However, the model is fine-tuned explicitly via RL to enhance its multi-turn tool interaction capabilities.
\end{enumerate}

We focus on Web Search and Code Interpreter as they represent two fundamental paradigms of agentic tool use. Evidence tools, exemplified by Web Search, are characterized by open-ended, stochastic outputs containing noisy, unstructured information. These properties are shared by other information-retrieval tools such as RAG and API queries. Conversely, verification tools, exemplified by Code Interpreter, provide deterministic, structured feedback that facilitates logical grounding. These properties are shared by other execution-based tools such as calculators, SQL, and symbolic solvers. Most other tools used by agents fall within the spectrum defined by these two paradigms. We evaluate three configurations across two representative tool categories using \textit{Qwen2.5-3B-Instruct} \cite{qwen2025qwen25technicalreport} as the backbone model:

\noindent \textbf{Evidence Tools (Web Search):} We evaluate open-domain question answering performance using the NQ \cite{kwiatkowski-etal-2019-natural} and HotpotQA \cite{yang-etal-2018-hotpotqa} datasets. These tasks require agents to retrieve external information from a noisy retrieval environment. For the RL-based variant, we adopt the training configurations of Search-R1 \cite{jin2025searchr1trainingllmsreason} using the VeRL framework \cite{sheng2024hybridflow}.

\noindent \textbf{Verification Tools (Code Interpreter):} We assess mathematical reasoning capabilities using the AIME2024/2025 \cite{aime} and MATH-500 \cite{lightman2023letsverifystepstep} benchmarks. These tasks allow agents to execute Python code and receive deterministic feedback. The RL-based model is implemented following the SimpleTIR \cite{xue2025simpletirendtoendreinforcementlearning} methodology for tool-integrated reasoning.

We measure the \textbf{M}ean \textbf{C}onfidence on \textbf{I}ncorrect \textbf{P}redictions (MCIP) on the intersectional wrong questions across the three settings to understand how additional tool use affects confidence, defined as $\mathrm{MCIP} = \frac{1}{|\mathcal{D}_{\text{wrong}}|} \sum_{i \in \mathcal{D}_{\text{wrong}}} \hat{p}_i$, where $\mathcal{D}_{\text{wrong}} = { i \mid y^*_i \neq y_i }$ denotes the set of incorrectly answered questions and $\hat{p}_i$ is the model's predicted confidence for its chosen answer on example $i$. Specifically, we use this metric to check whether agents show significantly different confidence patterns across various tool-use configurations. To ensure the power of our analysis, we additionally conducted \textit{t}-tests \cite{10.1093/biomet/6.1.1} between each configuration.

\subsection{Results}
\begin{table}[t]
    \centering
    \setlength{\tabcolsep}{6pt}
    \resizebox{\columnwidth}{!}{%
    \begin{tabular}{lllcc}
        \toprule
        Domain & Dataset & Setting & Acc.\,$\uparrow$ & MCIP\,$\downarrow$ \\
        \midrule
        \multirow{6}{*}{Search} 
            & \multirow{3}{*}{NQ} 
                & Direct prompting              & 15.5 & 0.879 \\
            &                                   & Prompting-based tool-use & 34.3 & 0.901 \\
            &                                   & RL-based tool-use        & 43.1 & 0.948 \\
            \cmidrule(lr){2-5}
            & \multirow{3}{*}{HotpotQA} 
                & Direct prompting              & 18.7 & 0.859 \\
            &                                   & Prompting-based tool-use & 22.4 & 0.911 \\
            &                                   & RL-based tool-use        & 27.4 & 0.967 \\
        \midrule
        \multirow{9}{*}{Code} 
            & \multirow{3}{*}{AIME2024} 
                & Direct prompting              & 12.2 & 0.968 \\
            &                                   & Prompting-based tool-use & 4.4 & 0.915 \\
            &                                   & RL-based tool-use        & 18.2 & 0.868 \\
            \cmidrule(lr){2-5}
            & \multirow{3}{*}{AIME2025} 
                & Direct prompting              & 15.7 & 0.968 \\
            &                                   & Prompting-based tool-use & 5.8 & 0.904 \\
            &                                   & RL-based tool-use        & 19.8 & 0.879 \\
            \cmidrule(lr){2-5}
            & \multirow{3}{*}{MATH-500} 
                & Direct prompting              & 63.4 & 0.971 \\
            &                                   & Prompting-based tool-use & 48.6 & 0.913 \\
            &                                   & RL-based tool-use        & 77.0 & 0.890 \\
        \bottomrule
    \end{tabular}
    }
    \caption{Pilot study results across tool-use configurations and domains using \textit{Qwen2.5-3B-Instruct} as the backbone. Accuracy (Acc.) is reported for all tasks. Lower MCIP indicates fewer overconfidence issues.}
    \label{tab:pilot_study_results}
    \vspace{-1em}
\end{table}

The empirical results of our pilot study, detailed in \autoref{tab:pilot_study_results}, reveal a critical \textbf{confidence dichotomy} in agent behavior. First, consistent with observations by \citet{wei2025browsecompsimplechallengingbenchmark}, integrating evidence tools (i.e., web search) leads to a marked deterioration in calibration. Regardless of whether the agent employs prompting-based strategies or RL-enhanced capabilities, the presence of search tools systematically exacerbates overconfidence. And our \textit{t}-tests suggest that the differences are statistically significant from direct prompting to RL-based tool-use. This suggests that the inherent noise and ambiguity of retrieved information predispose agents to inflated certainty. Conversely, verification tools (i.e., Code interpreters) yield a sharply contrasting dynamic. In this domain, tool usage mitigates overconfidence, with RL-based agents achieving the lowest MCIP (statistically significant in the reverse order). These divergent outcomes challenge the assumption that tool use exerts a uniform influence on agent confidence. Instead, we identify a clear heterogeneity driven by tool type: evidence tools, which introduce external information with inherent noise, interfere with confidence estimation, whereas verification tools, which provide deterministic feedback, ground the reasoning process and temper unwarranted certainty. This dichotomy motivates our development of calibration-aware training strategies for evidence tool scenarios.

\section{Calibration Agentic RL (CAR)}
As demonstrated in our pilot study, agents operating with evidence tools are systematically prone to severe overconfidence due to the inherent noise in retrieved information. To mitigate this challenge in multi-turn agentic tasks, we introduce the \textbf{Calibration Agentic RL (CAR)} framework. This design is specifically engineered to enhance the agent's ability to provide reliable confidence estimates alongside its tool-use actions.

\subsection{Experimental Setup}
\subsubsection{Training Details}
We first focus on evidence tool scenarios where tool use exacerbates miscalibration, leveraging the Search-R1 framework \cite{jin2025searchr1trainingllmsreason} as our primary testbed.
For the retrieval component, we directly adopt the established local search engine configuration utilized within Search-R1. This setup employs the 2018 Wikipedia dump \cite{karpukhin2020dense} as the knowledge source, coupled with E5 \cite{wang2024textembeddingsweaklysupervisedcontrastive} as the dense retriever. Furthermore, to optimize the policy of our tool-use agents, we employ Group Relative Policy Optimization (GRPO) \cite{shao2024deepseekmathpushinglimitsmathematical} as the foundational RL algorithm, and the mixture of NQ \cite{kwiatkowski-etal-2019-natural} and HotpotQA \cite{yang-etal-2018-hotpotqa} as our training data.

We employ a suite of instruction-tuned LLMs as our policy networks: \textit{Qwen2.5-3B-Instruct}, \textit{Qwen2.5-7B-Instruct}, and \textit{Qwen3-4B-Instruct-2507}. The selection of instruction-tuned variants is driven by two factors. First, these models exhibit superior adherence to complex instructions, facilitating more reliable verbalized confidence reporting. Second, empirical evidence from the original Search-R1 study indicates negligible performance disparities between base and instruction-tuned architectures in this context. 

\subsubsection{Reward Design}
\paragraph{Baseline Methods} We evaluate CAR against three baseline methodologies: (1) \textbf{Vanilla Search-R1}: the reward architecture proposed by \citet{jin2025searchr1trainingllmsreason}, which utilizes an exact match (EM) outcome reward alongside a structural reward to enforce adherence to the reasoning-action-observation chain; (2) \textbf{Temperature Scaling}: a verified post-hoc calibration method \cite{ece_and_temp_scaling} applied to the vanilla Search-R1 baseline with temperature fixed at 1.5. Superior performance over this baseline would indicate that the model has internalized genuine calibration capabilities rather than merely adjusting surface-level probabilities; and (3) \textbf{MASH}: \citet{gul2025paypersearchmodelsabstentionmodels} introduced a penalty mechanism for excessive search tool usage that fosters robust abstention behavior. Given the correlation between selective abstention and improved calibration \cite{kirichenko2025abstentionbenchreasoningllmsfail, song-etal-2025-hallucination}, we include this as a comparative baseline.

\paragraph{Proposed Reward Architecture} Following \citet{damani2025binaryrewardstraininglms}, we augment the system prompt to require agents to output a numerical confidence score (ranging from 0 to 100) within \texttt{<confidence>} XML tags. Our reward design comprises two components:

\paragraph{(1) Extended Format Reward} To ensure structural integrity, we extend the standard Search-R1 format constraints. While the original formulation validates the logical ordering of reasoning, action, and observation, our design additionally mandates the presence of the confidence tag. Consequently, the boolean reward function $f_{format}(y)$ returns a value of $True$ only when all structural requirements, including the confidence encapsulation, are strictly satisfied.

\paragraph{(2) Calibration-motivated Outcome Reward} We reward both the accuracy of the final answer and the expressed confidence. Given the gold answer $y$, predicted answer $y^*$, and verbalized confidence $q$, we experiment with two formulations:

\noindent \textbf{Weighted Brier Score Reward.} Following \citet{damani2025binaryrewardstraininglms}, we use the Brier score \cite{glenn1950verification} to form a combined reward:
\begin{equation}
\label{eq:rrlcf}
\begin{split}
    \rrlcf(y, q, y^*) &= \mathbbm{1}_{y\equiv y^*} - \lambda (q - \mathbbm{1}_{y \equiv y^*})^2.
\end{split}
\end{equation}
When $\lambda = 1$, this reduces to the RLCR formulation. However, in this setting, the lowest reward for a correct attempt equals the highest reward for an incorrect one, which may make learning sensitive to the training data distribution. To restore a positive incentive margin for correctness, we experiment with a weighting coefficient $\lambda = 1/3$ on the Brier term.

\noindent \textbf{Margin-Separated Calibration Reward (MSCR).} To address the optimization instability caused by incentive overlap in Brier scores, we propose MSCR. This formulation decouples calibration terms for correct and incorrect predictions to guarantee a strict reward margin:
\begin{equation}
\label{eq:rsym}
\begin{split}
    R_{\text{MSCR}}(y, q, y^*) &= \mathbbm{1}_{y \equiv y^*} \left[ 1 + \beta_1 (1 - (1-q)^2) \right] \\
    &\quad - \mathbbm{1}_{y \not\equiv y^*} \left[ \beta_2 q^2 \right],
\end{split}
\end{equation}
where $\beta_1, \beta_2 > 0$ control the calibration magnitude. This formulation enforces strict separation: correct answers receive a base reward of at least 1 (even with $q=0$), while incorrect answers receive at most 0 (at $q=0$) and incur penalties for false confidence. This eliminates the ``safe failure'' loophole, ensuring that the least confident correct answer strictly outperforms the most ``honest'' incorrect answer.

\paragraph{Unified Reward Function} The total reward combines format constraints with calibration-aware scoring. Denoting the model output as $\tau$, the chosen calibration function as $\mathcal{R}_{\text{cal}}(y, q, y^*)$ and assigning a penalty $\lambda_f$ for format violations:
\begin{equation}
\label{eq:reward_summary}
\small
r_{\phi}(x, \tau) = \begin{cases}
    \mathcal{R}_{\text{cal}}(y, q, y^*) & \text{if } f_{\text{format}}(\tau) = \text{True}, \\
    \mathcal{R}_{\text{cal}}(y, q, y^*) - \lambda_f & \text{otherwise}.
\end{cases}
\end{equation}
\noindent In the \textit{otherwise} case, if $q$ cannot be extracted, we fall back to a minimal calibration reward (treating $q$ as a default value) to maintain the correctness gradient while penalizing format errors via $\lambda_f$.

\subsubsection{Evaluation Details}
We evaluate our trained agents on the following benchmarks: (1) the validation sets of NQ and HotpotQA, serving as in-distribution (ID) datasets; and (2) SimpleQA-verified \cite{haas2025simpleqaverifiedreliablefactuality}, a curated subset of SimpleQA \cite{wei2024measuringshortformfactualitylarge} comprising 1,000 rigorously filtered questions for out-of-distribution (OOD) assessment. The retrieval corpus remains the 2018 Wikipedia dump for all local-retriever evaluations.
We employ four metrics for comprehensive analysis: (1) Accuracy (Acc.), to verify that calibration improvements do not come at the cost of task performance; (2) Expected Calibration Error (ECE), the canonical metric for confidence-accuracy alignment, calculated using a 10-bin scheme; (3) Brier Score, which captures both calibration and refinement as the squared difference between confidence and correctness; and (4) AUROC, to assess whether agents can reliably distinguish correct from incorrect predictions via confidence ranking.

\section{Results}

\begin{table*}[t]
\centering
\small
\setlength{\tabcolsep}{3.5pt}
\resizebox{\textwidth}{!}{
\begin{tabular}{ll|cccc|cccc|cccc}
\toprule
\multirow{2}{*}{Model} & \multirow{2}{*}{Strategy} &
\multicolumn{4}{c|}{NQ (ID)} &
\multicolumn{4}{c|}{HotpotQA (ID)} &
\multicolumn{4}{c}{SimpleQA-verified (OOD)} \\
& &
Acc$\uparrow$ & ECE$\downarrow$ & Brier$\downarrow$ & AUROC$\uparrow$ &
Acc$\uparrow$ & ECE$\downarrow$ & Brier$\downarrow$ & AUROC$\uparrow$ &
Acc$\uparrow$ & ECE$\downarrow$ & Brier$\downarrow$ & AUROC$\uparrow$ \\
\midrule
% -------------------- Qwen2.5-3B --------------------
\multirow{6}{*}{Qwen2.5-3B}
& Vanilla Search-R1             & 43.1 & 0.528 & 0.519 & 0.600 & 27.4 & 0.699 & 0.686 & 0.599 & 34.8 & 0.610 & 0.587 & 0.702 \\
& Temperature Scaling      & 43.1 & 0.500 & 0.489 & 0.600 & 27.4 & 0.674 & 0.651 & 0.599 & 34.8 & 0.583 & 0.549 & 0.702 \\
& MASH                          & 43.4 & 0.479 & 0.470 & 0.612 & 28.9 & 0.656 & 0.631 & 0.598 & 35.1 & 0.589 & 0.560 & 0.696 \\
\cdashline{2-14}
& CAR (Weighted Brier, $\lambda{=}1$)     & 22.3 & 0.091 & 0.091 & 0.941 & 24.3 & 0.148 & 0.148 & 0.902 & 21.2 & 0.027 & 0.027 & 0.983 \\
& CAR (Weighted Brier, $\lambda{=}1/3$)   & 44.5 & 0.307 & 0.313 & 0.677 & 28.5 & 0.329 & 0.329 & 0.651 & 35.8 & 0.203 & 0.203 & 0.740 \\
& CAR (MSCR)               & 45.5 & 0.303 & 0.303 & 0.699 & 29.2 & 0.286 & 0.289 & 0.688 & 36.6 & 0.192 & 0.191 & 0.763 \\
\midrule
% -------------------- Qwen2.5-7B --------------------
\multirow{6}{*}{Qwen2.5-7B}
& Vanilla Search-R1             & 61.1 & 0.363 & 0.367 & 0.563 & 54.8 & 0.424 & 0.423 & 0.551 & 40.7 & 0.441 & 0.398 & 0.776 \\
& Temperature Scaling      & 61.1 & 0.311 & 0.330 & 0.563 & 54.8 & 0.371 & 0.380 & 0.551 & 40.7 & 0.385 & 0.347 & 0.776 \\
& MASH                          & 67.0 & 0.309 & 0.315 & 0.543 & 55.6 & 0.421 & 0.422 & 0.521 & 41.5 & 0.465 & 0.420 & 0.775 \\
\cdashline{2-14}
& CAR (Weighted Brier, $\lambda{=}1$)     & 65.2 & 0.221 & 0.242 & 0.693 & 48.7 & 0.330 & 0.358 & 0.583 & 24.4 & 0.050 & 0.053 & 0.940 \\
& CAR (Weighted Brier, $\lambda{=}1/3$)   & 67.7 & 0.281 & 0.293 & 0.629 & 52.8 & 0.368 & 0.379 & 0.600 & 40.5 & 0.177 & 0.176 & 0.798 \\
& CAR (MSCR)               & 69.3 & 0.238 & 0.255 & 0.637 & 56.8 & 0.326 & 0.348 & 0.641 & 40.9 & 0.150 & 0.156 & 0.837 \\
\midrule
% -------------------- Qwen3-4B --------------------
\multirow{6}{*}{Qwen3-4B}
& Vanilla Search-R1             & 45.7 & 0.452 & 0.438 & 0.634 & 46.6 & 0.408 & 0.391 & 0.671 & 42.2 & 0.287 & 0.210 & 0.874 \\
& Temperature Scaling      & 45.7 & 0.380 & 0.377 & 0.634 & 46.6 & 0.343 & 0.340 & 0.671 & 42.2 & 0.254 & 0.198 & 0.874 \\
& MASH                          & 36.3 & 0.543 & 0.514 & 0.645 & 35.5 & 0.498 & 0.457 & 0.695 & 37.4 & 0.299 & 0.213 & 0.859 \\
\cdashline{2-14}
& CAR (Weighted Brier, $\lambda{=}1$)     & 28.5 & 0.115 & 0.120 & 0.918 & 36.6 & 0.184 & 0.186 & 0.856 & 33.1 & 0.036 & 0.033 & 0.973 \\
& CAR (Weighted Brier, $\lambda{=}1/3$)   & 44.2 & 0.274 & 0.274 & 0.718 & 45.0 & 0.281 & 0.281 & 0.727 & 40.1 & 0.127 & 0.127 & 0.912 \\
& CAR (MSCR)               & 45.5 & 0.272 & 0.272 & 0.724 & 45.9 & 0.269 & 0.270 & 0.740 & 41.8 & 0.106 & 0.106 & 0.929 \\
\bottomrule
\end{tabular}}
\caption{Main results organized by backbone model. Dashed lines separate baselines from CAR variants.}
\label{tab:main_results_by_model}
\vspace{-1em}
\end{table*}

\subsection{General Results on Search Agents}
A comprehensive summary of our experimental findings is presented in \autoref{tab:main_results_by_model}. The results demonstrate the robust effectiveness of CAR: across all three backbone models with different sizes, we consistently observe substantial ECE reductions compared to baseline methods. This improvement is consistent across both in-distribution (ID) and out-of-distribution (OOD) settings, with ECE relative reductions of up to 68\% through explicit calibration-aware RL training. Crucially, under our optimal configuration (MSCR), agents maintain accuracy levels competitive with reward structures that strictly incentivize correctness, confirming that our design successfully balances calibration and task performance. Furthermore, our analyses of AUROC and temperature scaling suggest that CAR achieves these gains not through mere rescaling of confidence scores, but by inducing more nuanced confidence reasoning. This is evidenced by AUROC relative improvements of up to 17\% in our best setting, confirming that the model has genuinely improved its ability to distinguish correct from incorrect outputs.

This improved reasoning capability translates into robust generalization, as evidenced by the comparison between ID and OOD settings. Our results indicate that CAR engenders a fundamental understanding of confidence rather than a superficial alignment with in-distribution data. Specifically, on the SimpleQA-verified dataset, agents trained via CAR exhibit marked calibration improvements. This finding suggests that the calibration mechanisms learned in search scenarios are not artifacts of the training set but instead transferable skills that generalize reliably to less familiar queries.

A comparative analysis of the three CAR configurations reveals the critical role of reward gap magnitude. The weighted Brier score with $\lambda{=}1$ (i.e., vanilla RLCR) yields the lowest ECE but suffers from significant accuracy degradation, indicating reward hacking behavior. In contrast, MSCR achieves a superior accuracy-calibration trade-off: across most settings, it attains higher accuracy than the weighted Brier variant with $\lambda{=}1/3$ while simultaneously improving calibration. These results suggest that strict reward separation is essential for robust calibration training, a point we discuss further in Section~\ref{sec:discussion}.

\subsection{Tool generalization}
Our primary experiments employed a simulated retriever environment with a static Wikipedia dump. However, real-world deployment poses additional challenges: commercial API-based retrievers often exhibit stochastic behavior and return noisy or extraneous information. In this section, we investigate whether the calibration capabilities learned in controlled settings transfer to these more challenging, API-driven environments.

\paragraph{Setup} We use the Serper API as our retrieval backbone and evaluate both vanilla Search-R1 and CAR (MSCR) on the SimpleQA-verified benchmark.

\begin{table}[t]
\centering
\small
\setlength{\tabcolsep}{5pt}
\resizebox{\columnwidth}{!}{
\begin{tabular}{ll|cccc}
\toprule
\multirow{2}{*}{Model} & \multirow{2}{*}{Method} &
\multicolumn{4}{c}{SimpleQA-verified (Serper API)} \\
& & Acc$\uparrow$ & ECE$\downarrow$ & Brier$\downarrow$ & AUROC$\uparrow$ \\
\midrule
\multirow{2}{*}{Qwen2.5-3B}
& Vanilla S-R1 & 76.18 & 0.213 & 0.219 & 0.659 \\
\cdashline{2-6}
& CAR (MSCR) & 76.18 & 0.175 & 0.175 & 0.823 \\
\midrule
\multirow{2}{*}{Qwen2.5-7B}
& Vanilla S-R1 & 70.28 & 0.204 & 0.204 & 0.831 \\
\cdashline{2-6}
& CAR (MSCR) & 71.01 & 0.176 & 0.180 & 0.790 \\
\midrule
\multirow{2}{*}{Qwen3-4B}
& Vanilla S-R1 & 85.27 & 0.140 & 0.140 & 0.825 \\
\cdashline{2-6}
& CAR (MSCR) & 84.97 & 0.034 & 0.073 & 0.765 \\
\bottomrule
\end{tabular}}
\caption{Tool generalization under a noisy API-driven retriever (Serper API). We evaluate Vanilla Search-R1 and CAR (MSCR) on SimpleQA-verified.}
\label{tab:tool_generalization_serper_r1_vs_car}
\end{table}

\begin{table*}[t]
\centering
\small
\setlength{\tabcolsep}{4pt}
\resizebox{\textwidth}{!}{
\begin{tabular}{ll|cccc|cccc|cccc}
\toprule
\multirow{2}{*}{Model} & \multirow{2}{*}{Method} &
\multicolumn{4}{c|}{AIME2024} &
\multicolumn{4}{c|}{AIME2025} &
\multicolumn{4}{c}{MATH-500} \\
& 
& Acc$\uparrow$ & ECE$\downarrow$ & Brier$\downarrow$ & AUROC$\uparrow$
& Acc$\uparrow$ & ECE$\downarrow$ & Brier$\downarrow$ & AUROC$\uparrow$
& Acc$\uparrow$ & ECE$\downarrow$ & Brier$\downarrow$ & AUROC$\uparrow$ \\
\midrule
\multirow{2}{*}{Qwen2.5-3B}
& Vanilla SimpleTIR  & 18.2 & 0.692 & 0.630 & 0.489 & 19.8 & 0.687 & 0.632 & 0.498 & 77.0 & 0.151 & 0.193 & 0.622 \\
\cdashline{2-14}
& CAR (MSCR)    & 20.8 & 0.573 & 0.485 & 0.548 & 21.1 & 0.519 & 0.410 & 0.695 & 76.9 & 0.057 & 0.168 & 0.636 \\
\bottomrule
\end{tabular}}
\caption{Results on mathematical reasoning benchmarks of tool-integrated reasoning agents. }
\label{tab:simpletir_results}
\vspace{-1em}
\end{table*}

As shown in \autoref{tab:tool_generalization_serper_r1_vs_car}, our trained agents achieve superior calibration compared to the vanilla baseline while maintaining competitive accuracy. These results confirm that the calibration capabilities acquired in simulated environments are not brittle but transfer robustly to the stochastic and noisy conditions of real-world API interactions.

\subsection{Tool-integrated Reasoning} 
Building on our pilot study, we extend evaluation to Tool-integrated Reasoning (TIR), where agents leverage code interpreters to solve mathematical problems. This extension allows us to examine how CAR performs with verification tools, which our pilot study identified as inducing calibration dynamics different from those of evidence tools.

\paragraph{Setup} We utilize the SimpleTIR \cite{xue2025simpletirendtoendreinforcementlearning} framework with \textit{Qwen2.5-3B-Instruct} as the backbone model. We compare two configurations: vanilla SimpleTIR as the baseline and our MSCR design. For evaluation, we adopt the AIME2024/2025 \cite{aime} and MATH-500 \cite{lightman2023letsverifystepstep} benchmarks, utilizing E2B\footnote{https://e2b.dev/} as the code execution sandbox.

\paragraph{Results} \autoref{tab:simpletir_results} presents the quantitative results. Consistent with our findings in the search domain, CAR yields robust calibration improvements for TIR agents, with significant reductions in ECE and Brier scores alongside increases in AUROC. These gains persist across all evaluated benchmarks, confirming the generalizability of our reward formulation.

However, examining absolute performance reveals an important nuance. Despite these improvements, ECE metrics for TIR agents remain elevated compared to both pure reasoning models \cite{zeng-etal-2025-thinking} and our search-based agents. Furthermore, calibration efficacy correlates with task complexity: agents exhibit substantially lower ECE on MATH-500 than on the more challenging AIME benchmarks. These observations align with our pilot study hypothesis that verification tools, while providing grounding through deterministic feedback, still exhibit calibration dynamics that depend on task difficulty. We conclude that while explicit calibration rewards provide a necessary corrective signal, ultimate calibration performance in TIR settings remains bounded by the model's intrinsic reasoning capabilities.

\section{Discussion}
\label{sec:discussion}
Our findings reveal that tool use introduces systematic yet heterogeneous effects on agent calibration, challenging the implicit assumption that tool augmentation uniformly improves or degrades reliability. In this section, we discuss the broader implications of these findings and examine the mechanisms underlying the observed confidence dichotomy.

\paragraph{From Static Elicitation to Tool-Modulated Dynamics.} The transition from static LLMs to autonomous agents necessitates a fundamental re-evaluation of verbalized calibration paradigms. While extensive literature establishes that models can accurately express uncertainty in single-turn QA \cite{tian-etal-2023-just} or standard Chain-of-Thought reasoning \cite{xiong2024llmsexpressuncertaintyempirical, zeng-etal-2025-thinking}, our analysis reveals that tool integration introduces a non-trivial \emph{heterogeneity} that disrupts this alignment. Specifically, we contextualize the severe miscalibration observed in recent web-browsing agents \cite{wei2025browsecompsimplechallengingbenchmark, zhou2025browsecompzhbenchmarkingwebbrowsing} not merely as a general capability failure, but as a symptom of the evidence tool dynamic, where stochastic retrieval artificially inflates internal certainty. This stands in sharp contrast to verification scenarios, where deterministic feedback provides the grounding often assumed but absent in open-ended search. These findings suggest that calibration research in agentic settings must account for tool-type-specific dynamics rather than treating tool use as a monolithic phenomenon.

\paragraph{Why Do Evidence Tools Induce Overconfidence?}
Our pilot study indicates that evidence tools systematically amplify overconfidence, but the underlying mechanism warrants further examination. We attribute this phenomenon to a fundamental asymmetry in feedback signals. Verification tools such as code interpreters provide explicit execution feedback: syntax errors, runtime exceptions, and type mismatches offer clear signals that something is wrong. While successful execution does not guarantee correctness, as logical errors may still produce plausible but incorrect outputs, these tools nonetheless provide \emph{partial grounding} via observable failure modes. In contrast, evidence tools offer little answer-level correctness feedback. A web search typically returns results, regardless of whether they are relevant, accurate, or sufficient to answer the query. The absence of negative feedback leads agents to conflate the \emph{presence} of retrieved information with the \emph{correctness} of their answer. This effect is compounded by a form of false certainty induced by the retrieval action itself. Having performed an explicit information-seeking step, the agent treats it as ``due diligence,'' even when the retrieved content is noisy or misleading. Retrieved passages often contain surface-level lexical overlap with the query, which the agent mistakes for genuine evidential support.

\paragraph{Extending RLCR to Tool-Use Agents via MSCR.}
Our approach builds on calibration-motivated RL, particularly RLCR \cite{damani2025binaryrewardstraininglms}, which shows that optimizing binary correctness rewards can degrade calibration by encouraging guessing and proposes incorporating calibration terms into the training objective. We extend this principle to the agentic setting. Tool-use agents face additional exogenous noise and expanded action spaces, thereby enlarging the space of degenerate solutions in which confidence becomes uninformative or strategically manipulated. Concretely, we find that reward overlap between correct and incorrect trajectories makes learning sensitive to data difficulty and can incentivize confidence collapse. Our proposed MSCR addresses this by enforcing strict separation between reward landscapes for correct and incorrect trajectories, preserving a correctness margin while still shaping confidence within each region. Empirically, this design improves calibration and failure discrimination beyond what simple rescaling would achieve, and generalizes to noisy retrieval environments, suggesting that calibration-motivated RL remains effective in multi-step tool use provided that incentive separation is maintained.

\section{Conclusion}
In this work, we systematically investigated the calibration dynamics of tool-use agents. Our pilot study revealed a fundamental confidence dichotomy: while verification tools provide deterministic feedback that grounds reasoning, evidence tools introduce stochastic noise that systematically induces overconfidence. To address the miscalibration in tool-use agents, we proposed the Calibration Agentic RL (CAR) framework, incorporating a novel Margin-Separated Calibration Reward (MSCR) that strictly separates incentives for correct and incorrect predictions. Extensive experiments demonstrate that CAR significantly reduces calibration error while maintaining competitive task performance, with robust generalization from local simulation to noisy, real-world API environments. Our findings underscore the necessity of tool-specific calibration strategies and establish a foundation for building self-aware agents capable of reliably communicating uncertainty in high-stakes deployments.

\section*{Limitations}
In this work, we studied the confidence dichotomy between evidence and verification tools in controlled agentic settings and proposed the Calibration Agentic RL framework to address miscalibration in evidence-tool scenarios. One key limitation is that our experiments focus on models with 3B to 7B parameters due to computational constraints. While we observe consistent patterns across three backbone architectures, it remains unclear how this phenomenon evolves with scale. Furthermore, our evaluation primarily focuses on short-answer question answering and mathematical reasoning, where correctness is well-defined. Calibration behavior in more open-ended generation scenarios, such as long-form report writing or multi-step autonomous planning, may involve underspecified correctness signals or more delayed feedback loops that our current framework does not address. We leave these explorations to future work.

\bibliography{custom}

\end{document}